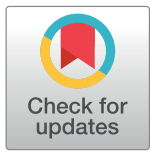

RESEARCH ARTICLE

# CIDER: Context-sensitive polarity measurement for short-form text

**James C. Young***, **Rudy Arthur**, **Hywel T. P. Williams**

Computer Science, Innovation Centre, University of Exeter, Exeter, United Kingdom

* jcy204@exeter.ac.uk

## Abstract

Researchers commonly perform sentiment analysis on large collections of short texts like tweets, Reddit posts or newspaper headlines that are all focused on a specific topic, theme or event. Usually, general-purpose sentiment analysis methods are used. These perform well on average but miss the variation in meaning that happens across different contexts, for example, the word "active" has a very different intention and valence in the phrase "active lifestyle" versus "active volcano". This work presents a new approach, CIDER (Context Informed Dictionary and sEmantic Reasoner), which performs context-sensitive linguistic analysis, where the valence of sentiment-laden terms is inferred from the whole corpus before being used to score the individual texts. In this paper, we detail the CIDER algorithm and demonstrate that it outperforms state-of-the-art generalist unsupervised sentiment analysis techniques on a large collection of tweets about the weather. CIDER is also applicable to alternative (non-sentiment) linguistic scales. A case study on gender in the UK is presented, with the identification of highly gendered and sentiment-laden days. We have made our implementation of CIDER available as a Python package: https://pypi.org/project/ciderpolarity/.

**Data Availability Statement:** All relevant data are within the paper and its Supporting information files.

**Funding:** H.T.P.W. acknowledges funding from UK Natural Environment Research Council (NE/P017436/1). J.C.Y. is funded by a PhD studentship from the UK Engineering and Physical Sciences Research Council. No funding bodies had any influence over the content of this report.

**Competing interests:** The authors have declared that no competing interests exist.

## 1 Introduction

Many words change their meaning and sentiment depending on the context in which they appear. In a discussion of health, "active" is a positive term, in a discussion of volcanoes it is not. Sarcasm e.g. "I would *love* to see that", can completely switch the sentiment of a phrase. Specific communities or cultures can also use words in different ways from their standard meaning, e.g. in the UK "clown" is a common insult (or in online text, the clown emoji). Automated sentiment detection methods are usually designed to work on any corpus of text, ignoring context [1]. Context dependence of language can lead to mislabelling and misquantification of meaning [2].

While sentiment analysis is perhaps the most extensively studied form of polarity assignment in natural language processing (NLP), it is not the only dimension along which text can be classified. Other works, such as Lucy et al. [3] and Zhao et al. [4], have explored scoring words on gender and morality dimensions, respectively. Since context is important in any application where text is placed on a scale, the gender associated with a term will depend on any number of factors e.g. the perceived "maleness" of an activity like football will likely depend on the popularity of the men's or women's game which can vary over time and place. Bolukbasi et al. [5] discuss potentially counter-intuitive cases that occur like the association of maleness with nursing arising from the prevalence of the phrase "male nurse". There is therefore a need for automated methods that can assign polarity scores to words in a context-sensitive manner. This paper aims to improve sentiment analysis, and any other polarity assignment task, by providing a straightforward way to incorporate domain-specific contextual information. Particularly, we focus on a common use case in (social) media analysis involving a large number of relatively short texts on specific topics, such as tweets, Reddit posts, and news headlines, and we want to tag each item on a scale from negative (-1) to positive (+1). These short texts present unique challenges for linguistic analysis due to their brevity, informal language, and frequent use of slang and emojis, which can alter the intended meaning. Additionally, they often lack the context provided in longer texts, making it difficult to discern the sentiment accurately. However, their prevalence and real-time nature make them invaluable sources of data for understanding and analysing current trends and public opinions [6].

General-purpose sentiment analysis approaches have often been used for extracting emotive posts on social media with research into tracking changes in public sentiment during extreme weather events [7–9], monitoring social unrest [10, 11], analysing discussions on climate change [12–14], and investigating terror attacks [15–17]. A popular and illustrative sentiment analysis model is VaderSentiment (VADER) [18], a rule-based sentiment classification algorithm that has been optimised for social media content. VADER uses a list of 7500 words (called a dictionary/lexicon) with manually assigned polarities (scores measuring positive or negative feeling) to measure sentiment. VADER also uses built-in mutation rules to handle negation, boosting, etc. (for instance, VADER assigns, "that is very BAD!" as more negative than, "that is bad") which greatly improves performance versus simple token counting. Although performing well on average, VADER can struggle when words are used outside of their most common context. This commonly occurs when discussing the weather. For example, VADER classifies both "Help me, there is a very strong storm!!" and "My house has been wrecked by an active volcano #alert" as statements of positive feeling due to its positive scoring of "help", "strong", "active", and "alert". One solution to this issue is to manually update the sentiment lexicon for each new domain. However, this approach is impractical, requiring significant human effort to understand contextual variations in each corpus, and may potentially become outdated, due to the rapid evolution of language on social media [19–21]. Another approach to resolving context-dependent sentiment is the supervised training of a novel classifier for a given domain [22–24]. Such methods can have high accuracy [2], however, they are costly to produce, requiring thousands of sample messages to be labelled manually, may not be robust against future evolution of language and can only be reliably used for a single application. Large language models (LLMs) [25] can be used; however, these are computationally expensive, may have limited specific domain knowledge, can suffer from hidden biases, and lack explainability.

To address this trade-off between high-accuracy high-cost methods and low-accuracy low-cost methods, we present a new sentiment analysis package called CIDER (Context Informed Dictionary and sEmantic Reasoner). CIDER requires minimal supervision and is automatically tuned to a particular domain or context. CIDER is based on combining the SocialSent algorithm [26] with VADER. SocialSent is a technique developed by Hamilton et al. [26] which can create domain-specific lexicons using a small set of positive and negative seed words as its only input. Our approach is to first construct a domain-specific lexicon using SocialSent by creating short lists of relevant seed words, and then filter and substitute this lexicon into VADER. This combines the ability of SocialSent to create a lexicon in a mostly unsupervised way with VADER's proven ability to handle sentiment in short texts and in particular, social media posts. VADER's boosting and negation rules can also be applied to non-sentiment scales (e.g. "really hot", "not cold") allowing us to go beyond word-level analysis for other polarity axes.

A notable example of polarity assignment on other scales is SemAxis [27] which provides a general framework for scoring *words* along arbitrary scales. Mathew et al. [28] demonstrates a different approach based on word embeddings with similar aims, but uses pre-trained word embeddings and so sacrifices domain specificity. Both An et al. [27] and Mathew et al. [28] score individual words rather than sentences. Analysis at the sentence level is important in many applications [29]. For example, we will discuss a corpus of weather-related tweets in the following, evaluating contrastive sentences like "yesterday was freezing, today is ridiculously hot!" only makes sense at the sentence level, rather than the word level. Even simple negation, "Today is not hot", could lead to errors in a naive word counting analysis.

Two case studies are presented in this paper. The first case study uses weather-related social media (Twitter) content, but we assert that our approach would be similarly useful in many other application areas where language adapts to context. To validate the improvement offered by our approach, we compare performance against eight other unsupervised sentiment analysis models. Results show that CIDER performs significantly better than all unsupervised models, decreasing the gap between the lighter-weight models and the more computationally and labour-intensive supervised models. The second case study demonstrates the use of CIDER for non-sentiment scales, creating a gender classifier which can be combined with the sentiment classifier, enabling multi-dimensional analysis of text, identifying days of high gender and sentiment intensity. Such multi-dimensional analysis can uncover patterns and correlations that single-factor studies might miss, offering insights into how gender interacts with emotional expression [30]. This can be particularly valuable in areas like targeted marketing, sociolinguistic research, and understanding public sentiment on gender-related issues, providing a richer, more layered understanding of online discourse [31].

To support our experimental findings, CIDER has been released on both GitHub (available here: https://github.com/jcy204/ciderPolarity), and PyPi (available here: https://pypi.org/project/ciderpolarity/, and through `pip install ciderPolarity`).

The structure of this paper is as follows. Section 2 (Methods) describes the CIDER methodology, starting with the underlying technique adopted from SocialSent, adaptations we have made to the algorithm for this use case, and how SocialSent was combined with VADER to infer sentiment. Section 3 (Experiment Design) presents the two datasets used. We then describe how CIDER was optimised in the two case studies. Section 4 (Results) presents the results from our validation experiments and some additional analyses. Section 5 (Discussion) gives some interpretation of the findings and offers some areas for future research.

## 2 Methods

Whilst both SocialSent and VADER are standalone tools, the following section highlights the modifications required to create CIDER, a single, easy-to-use, NLP pipeline. Additionally, we introduce an algorithm integrated into CIDER which has been designed to suggest potential seed words to further optimise its performance.

### 2.1 SocialSent

The learning phase of CIDER is based on the SocialSent algorithm [26]. For a comprehensive explanation, including justifications for each step, readers are referred to the original SocialSent paper. Below is a summary of the key steps involved in this process:

1. The corpus is first cleaned, removing punctuation/stopwords and ensuring all the text is in the same case, before being tokenised into unigrams.

2. A positive pointwise mutual information (PPMI) matrix is then constructed from the tokenised data. A PPMI matrix is used to compare the probability of word co-occurrence to word independence across the dataset.

3. Word vectors reflecting the relationships between words in the dataset are generated by taking the singular value decomposition (SVD) of the PPMI matrix.

4. A weighted lexical graph representing the semantic relationship between the SVD word vectors is then constructed. Each vector (word) is represented as a node, with each node connected to its K nearest word vectors (K = 25 in the original paper) in this semantic space. These neighbours are determined by finding the K closest word vectors based on cosine similarity. The edge weight between these vectors is defined as their respective cosine similarity. Cosine similarity is a standard metric in natural language processing [32–34] that measures the cosine of the angle between two vectors, representing how closely related two words or documents are in their meaning or content.

5. Label propagation using random walks from the location of a small number of manually provided positive and negative seed words within the graph is carried out, measuring each word's proximity to the positive and negative seeds independently.

6. The polarity for each word in the graph is then calculated using the word's average random walk distance to the positive seed words and the negative seed words respectively. The calculation for this is as follows:

$$word(i)_{polarity} = \frac{word(i)_{pos\_prox}}{word(i)_{pos\_prox} + word(i)_{neg\_prox}} \quad (1)$$

In the process of developing CIDER (available at https://pypi.org/project/ciderpolarity/) we have made a number of improvements to SocialSent:

1. Performance optimisation. The original SocialSent approach faced challenges with large datasets, leading to inefficiencies in memory usage and computation time. This was rectified by: (a) streaming the data rather than loading it all into local memory, and (b) parallelisation of calculations that previously occurred serially. Table 1 summarises the performance differences between the original and improved versions for multiple different dataset sizes.

2. Improvements for short-form text. Previous versions collated all individual documents (tweets/news articles/Reddit posts/etc.) into one large document and then applied a sliding window to generate the PPMI matrix. Here an option to treat documents individually has been added. Treating documents individually prevents polarities from potential confusion at document boundaries e.g. if the separate posts: "I love BTS", and "Yeah, I hate when people criticise them" were treated as one document this would put "hate" close to "BTS", contrary to the intended meaning.

Table 1. Comparing the performance of word valence calculations using SocialSent to CIDER.

| Datasets | SocialSent | | | CIDER | | |
|---|---|---|---|---|---|---|
| Rows | Size (MB) | RAM usage (MB) | Speed (s) | RAM usage (MB) | Speed (s) | Speedup |
| **3000** | 0.26 | 11.36 | 5.42 | 1.80 | 0.79 | 6.9x |
| **30,000** | 2.50 | 17.48 | 13.23 | 4.59 | 2.71 | 4.9x |
| **300,000** | 23.71 | 990.00 | 179.13 | 167.86 | 29.38 | 6.1x |
| **3,000,000** | 239.12 | 9889.50 | 1767.44 | 1960.60 | 232.93 | 7.6x |
| **30,000,000** | 2468.12 | 73,034.51 | 14,664.05 | 20,809.13 | 1790.87 | 8.2x |

For sufficiently large corpuses CIDER is approximately 8 times faster and uses significantly less RAM. Each result was calculated 20 times with the average presented.

https://doi.org/10.1371/journal.pone.0299490.t001

3. Seed word selection. The intention of seed words is for them to be both unambiguously polarised and have significant coverage within the corpus. To assist the user with identifying potential seed words, a function to generate custom seed words has been added (described in Section 2.3).

4. Parameter optimisation. Parameters in sentiment analysis can greatly influence the accuracy and reliability of results, and their optimal selection can be non-trivial. Using a grid search (discussed in section 3.2.2), we have optimised the parameter selection for sentiment analysis.

5. Refactored for readability, maintainability, and accessibility. The original SocialSent package, while powerful, was not user-friendly, limiting its applicability beyond computer science. CIDER addresses this by refactoring for improved readability and ease of use. Enhancements include streamlined code, better documentation, and a simplified interface, making it more accessible and easier to integrate into various workflows. These improvements make CIDER straightforward to install and use, extending its utility to a wider audience.

## 2.2 VADER

The VADER algorithm was largely retained in its original form to maintain its simplicity and explainability. Traditional sentiment analysis often categorises text on a linear scale from positive to negative. While this approach is effective for capturing strongly positive or negative sentiments, it fails to account for texts that are intensely both. For instance, the statement "I bloody hate the weather today, excited for the best weather tomorrow" is emotionally charged but would yield a neutral VADER score. By identifying highly emotive posts, we can more accurately distinguish truly neutral content and enable research into emotionally mixed content. We define emotive posts as pieces of text focused on conveying a person's feelings, characterised by expressive language and emotional tone, as opposed to neutral, fact-based descriptions.

In CIDER, we calculate an 'intensity' metric alongside the standard 'pos', 'neg', 'neu', and 'compound' VADER scores. Similar to the Emotional Variance Analysis (EVA) tool [35], which focuses on the variance of emotional expressions in texts, CIDER's approach also acknowledges the complexity of emotions beyond simple polarity. However, unlike EVA which calculates variance, CIDER derives intensity by first applying VADER's default mutation rules for boosting and negation to calculate word-level polarity scores, and then taking the absolute values of these scores. This array of absolute polarities is processed through the existing VADER pipeline, resulting in an 'intensity' score ranging from 0 (low intensity) to 1 (high intensity). The sentence "I bloody hate the weather today, excited for the best weather tomorrow" therefore has a neutral sentiment polarity due to the cancellation of the positive and negative parts in the sum, but high sentiment intensity because strong sentiment is being expressed. This approach maintains the transparency and ease of understanding that VADER promotes, allowing users to quickly grasp the strength of emotional expression in a text.

### 2.3 Seed word selection

CIDER requires two sets of opposing polarity seed words as input (e.g. positive/negative, hot/cold, male/female). These can be chosen manually or using a semi-automated approach. In larger corpora, the label propagation is robust to the choice of initial seed words due to the greater depth of the produced lexical graph. For smaller corpora (<30,000 rows), we found that a larger set of seed words that were both frequent and sufficiently polarised within the data produced higher-quality polarity lexicons. Hamilton et al. [26] presented a small selection of positive and negative seed words, however, for this investigation more were added. The seed words should be frequent/important in the dataset to enable the label propagation to reach a sufficient depth in the lexical graph, thus returning a sufficiently sized lexicon. The seed words should also be strongly polarised to enable a wide range of polarities with the resulting lexicon. To achieve this, the following was carried out:

1. The PPMI matrix for all words within the corpus investigated is calculated.

2. Two opposing small sets of unambiguously polarised words are then manually provided. The size of these sets can be adjusted according to the user's confidence in how clearly polarised the chosen words are on their investigated linguistic scale. For example, in this investigation, the following two sets were selected for sentiment:

$$\text{Set1} = [\text{``good'', ``love'', ``amazing''}]$$
$$\text{Set2} = [\text{``bad'', ``hate'', ``terrible''}]$$

3. Every word in the corpus is then ranked based on the difference between its average PPMI score to Set1, and its average PPMI score to Set2. This calculation is shown in Eq 2, where N1 and N2 represent the number of elements in Set1 and Set2 respectively.

$$\text{PPMI\_Distance}(i) = \frac{1}{N1}\sum_{w \in \text{Set1}} \text{PPMI}(w, i) - \frac{1}{N2}\sum_{w \in \text{Set2}} \text{PPMI}(w, i) \quad (2)$$

4. Words that have a high absolute PPMI_Distance and that were strongly polarised within the VADER lexicon are then returned as potential seed words. The VADER lexicon was selected as it provides a large, manually labelled, and filtered set of polarised words to choose from. If it is not a sentiment task, the returned seed words are just those with a high absolute PPMI_Distance.

The above methodology has been included in the CIDER package as a member function that can be used before running CIDER. It provides a structured approach for seed word selection in CIDER, yet it retains flexibility for user discretion. In situations where seed words are obvious or lack subjectivity, users can confidently choose them manually, streamlining the process and ensuring relevance to the specific domain of their corpus.

## 3 Experiment design

This section is split into three subsections. The first (Section 3.1) covers the two datasets, explaining how they were obtained and filtered. The second (Section 3.2), sets up the first experiment, investigating the use of CIDER to improve domain-specific sentiment analysis. The third (Section 3.3), covers how CIDER can be used for scales other than sentiment, with an investigation into gender.

### 3.1 Data collection

Twitter is a microblogging platform where users can post short-form messages (240 characters) to their followers [36]. With its global coverage, high volume of daily posts (500 million per day [37]), and accessible API, Twitter is a commonly used platform for NLP research. Recent API changes within Twitter have made the data less accessible, however, these methods work on any text (as shown through the relationship between Twitter and Telegram data demonstrated by Young et al. [38]). For this study, two datasets have been investigated. The first is a manually labelled weather Twitter dataset. The second is a geographic dataset from the UK in 2022. All data handling, manipulation, and analysis have been carried out to comply with Twitter's API terms of service [39]. Due to Twitter's distribution limit of 1,500,000 tweet IDs, a subset of tweet IDs this size was randomly selected from the original dataset for public sharing (available: https://github.com/jcy204/CIDER_Data/).

**3.1.1 Weather tweets.** To evaluate CIDER as a sentiment quantification tool, a validation dataset was required. For this, a dataset of 124,360 manually filtered weather tweets collected by Asiaee T. et al. [40] as part of the "Dialogue Earth" project was obtained. Each tweet in this collection has been manually annotated as 'Positive', 'Negative', or 'Neutral', with the average number of annotators being 5.1 (std dev: 0.9). Whilst the dataset is old (2012), its purpose is only to show that the VADER lexicon can be improved for a particular domain using CIDER rather than derive any specific conclusions about this data. Only tweets with a 100% annotator agreement were kept for this investigation.

The tweets covered various weather events and therefore to evaluate the ability of CIDER on multiple domains, the dataset was separated into three subsets, wind tweets, hot weather tweets, and cold weather tweets. As the tweets were already all relevant to weather events, simple keyword filtering sufficed for separating the datasets. The keywords used to filter the datasets are shown in Appendix A in S1 File. A manual evaluation of 100 tweets from each subset showed a high accuracy from filtering (wind: 98%, heat: 92%, cold: 94%). The number of tweets in each subset is also shown in Appendix A in S1 File.

**3.1.2 GeoUK 2022 tweets.** Whilst the weather validation dataset was filtered and collected for sentiment classification, an additional dataset was required to demonstrate the capability of CIDER beyond sentiment analysis. Using the Twitter API V2, every geolocated tweet in 2022 was collected. These tweets contained either automatically geotagged coordinates (depending on the user's phone permissions), or a manually tagged location within the 'place' attribute. These were then filtered to keep only tweets in the UK. Whilst this is only a sample of the true volume of tweets from the UK (typically geotagged tweets consist of ∼1% of total tweet volume [9]), the high volume of tweets (35,990,879 tweets) provided a sufficient overview of tweets from the UK. Due to data collection outages, only 318 days of tweets are present in the dataset.

The intention was to use a dataset reflective of the language used in the UK, therefore, the only filtering carried out was the removal of bot accounts. For this, a simple filter of removing all tweets where the user's tweet count was above 0.1% of the total dataset was carried out. This reduced the total number of tweets to 28,581,644. A manual inspection of the remaining tweets showed a high relevance for human-produced tweets across a broad spectrum of topics.

### 3.2 Evaluating CIDER for sentiment

To demonstrate the ability of CIDER to quantify sentiment in a context-sensitive way, we perform an evaluation study using weather-related content from Twitter. Weather is a good target for this, as sentiment often changes in different weather conditions [2, 9, 41, 42]. For example, sentiment about rain is very different during the winter than during a heatwave or drought. Similarly, words like "heat", "cool", "breeze", and "sun" can be very different depending on whether the temperature is perceived as being too high or too low. In the context of discussions on climate change, "hot weather" or "high temperatures" take on an especially negative meaning. There are also numerous examples of common words which have different implications, especially about disasters and natural hazards e.g. "active lifestyle:active volcano", "landslide victory:deadly landslide", "lightning fast:lightning strike" etc.

Tweets about the weather are usually about conditions experienced by the author or a response to (usually negative) news stories about serious weather events [38], and thus should contain many examples of context dependence and therefore provide a good test case where CIDER should outperform other general-purpose methods.

**3.2.1 Seed words.** After applying the semi-automated seed word selection method from Section 2.3 on the weather tweets dataset, the following seed words were obtained:

```
positive_seeds = ["lovely", "excellent", "fortunate", "pleasant", "delightful",
                  "perfect", "loved", "love", "loves", "good", "beautiful",
                  "great", "enjoy", "gorgeous", "awesome", "nice", "amazing",
                  "excited", "❤", "😊"]
negative_seeds = ["bad", "horrible", "hate", "damn",  "shit", "shitty", "fuck",
                  "hell", "wtf", "hated", "stupid", "terrible", "awful", "sad",
                  "crap", "crappy", "nasty", "worst",  "bitch", "hates", "☹"]
```

**3.2.2 Generating and filtering polarities.** Due to VADER's built-in negation rules ("I do not love this weather" = negative sentiment), tweets that contained a VADER negation term were excluded from the data used to train CIDER. This is to prevent language following negation terms from being influenced by the surrounding words in the tweet. For instance, the above example of "I do not love this weather", would be ignored to prevent the seed word 'love' from incorrectly increasing the positivity of 'weather'.

The VADER lexicon is bimodally distributed between -4 and 4 and thus the algorithm has been fine-tuned to perform best on a lexicon with a similar distribution and range. To linearly scale the generated CIDER polarities whilst preserving both the polarity skew and zero-centred mean, the following calculation is used:

$$scaled(i)_{polarities} = 4 * \frac{polarities(i)}{max(|polarities|)} \qquad (3)$$

To train a model which maximises the F1 score between predicted and true labels, a grid search over the following parameter groups was carried out:

P1). **Minimum Word Frequency**. Words that occur below this frequency are excluded from CIDER, preventing rare words from skewing polarities. If the number is too low it also drastically increases computation time. This parameter is easily tuned depending on the size of the dataset.

P2). **Maximum Word Frequency**. Excluding very common words prevents the dilution of polarities. This is carried out twice in CIDER and thus has two distinct values, P2a and P2b.

P3). **Nearest Neighbours**. This determines the number of neighbours for each node in the lexical graph.

P4). **Neutrality Filter**. CIDER identified many words as weakly polarised. This can worsen the sentiment classification. If a word's positive proximity AND negative proximity (as discussed in Eq 1) are both in the bottom P4% of their respective groups, then the word is deemed neutral and is removed from the lexicon. Eq 4 shows this filter, where a Neutral value of 1 implies the word is to be removed.

$$T_{\text{pos\_prox}} = \text{Percentile}([word(1)_{\text{pos\_prox}}, \ldots, word(n)_{\text{pos\_prox}}], \text{ P4})$$

$$T_{\text{neg\_prox}} = \text{Percentile}([word(1)_{\text{neg\_prox}}, \ldots, word(n)_{\text{neg\_prox}}], \text{ P4})$$

$$\text{Neutral} = \begin{cases} 1, & (word(i)_{\text{pos\_prox}} < T_{\text{pos\_prox}}) \text{ and } (word(i)_{\text{nex\_prox}} < T_{\text{neg\_prox}}) \\ 0, & \text{otherwise} \end{cases} \quad (4)$$

P5). **Polarised Filter**. This filter orders the lexicon by the difference between every word's positive and negative proximity. It then keeps the top and bottom P5%. The returned words are then substituted into the VADER lexicon. This is represented in Eq 5.

$$\begin{aligned}
\Delta(i) &= word(i)_{\text{pos\_prox}} - word(i)_{\text{neg\_prox}}, \\
\text{Positive\_Words} &= \{word(i) \mid \Delta(i) \geq \text{Percentile}(\Delta, 1 - P5)\}, \\
\text{Negative\_Words} &= \{word(i) \mid \Delta(i) \leq \text{Percentile}(\Delta, 0 + P5)\}
\end{aligned} \quad (5)$$

P6). **Classification Filter**. To convert the numerical linear output of CIDER as a distinct categorical label ("Positive", "Negative", "Neutral"), polarity boundaries are calculated using P6.

The final parameter values are as follows: P1 = 16, P2a = 0.3, P2b = 0.4, P3 = 20, P4 = 0.55, P5 = 0.13, P6 is shown in Appendix B in S1 File.

After filtering, each dataset's custom CIDER classifier was applied to its respective tweets. Apart from the VADER lexicon values excluded through threshold P4, the remainder of the VADER lexicon was kept. These parameters have been encoded into the CIDER library, with the option to fine-tune them if desired.

**3.2.3 Comparison methods.** To evaluate CIDER beyond a comparison just to VADER, the CIDER and default VADER algorithms were compared to seven more unsupervised sentiment analysis techniques. In a comparison study of twenty-four different approaches by Ribeiro et al. [43], Umigon [44], LIWC15 [45], VADER [18], and AFINN [46] performed the best on social media data and have been included. In a second, similar study by Zimbra et al. [29], Sentiment140 [47] and SentiStrength [48] were the best-performing general-purpose sentiment analysis algorithms on tweets and have been included. The final two algorithms included in the comparison were TextBlob [49] due to its prevalence in Twitter studies [50–52], and the updated LIWC22 approach [53]. Specifics regarding how each unsupervised algorithm is converted into a positive, negative, or neutral score are covered in Appendix B in S1 File.

In the present study, we have chosen to focus exclusively on unsupervised sentiment analysis methods, despite the availability of labeled data for model testing. The emphasis on unsupervised techniques aims to demonstrate their utility in situations where labelled datasets are either scarce or expensive to produce.

### 3.3 Evaluating CIDER on alternative polarity dimensions

The applicability of CIDER is not limited to sentiment, with multiple possible scales/axes for investigation. As a demonstration and case study, we focus on the representation of gender on Twitter. Gender on social media is a popular area of research, with differences in male and female communication styles existing [54, 55]. Starting with the pioneering work of Bolukbasi et al. [5] on gender bias in word embeddings, much work has been done to study and understand both how gender is a factor in written communication and how gender biases are reflected in NLP tools and analyses; see the review by Sun et al. [56]. Our aim is not to understand gendered communication *per se*, but to show how CIDER works with a scale other than sentiment and how it can be useful in this active research area. At the same time we recognise, following Devinney et al. [57], that CIDER is still a coarse tool and, like almost all studies of gender in NLP, we are limited in our ability to differentiate between biological, social and linguistic gender categories. In the words of Devinney et al. [57] we use a "cisnormative folk model" of gender and rely on future work which is aimed at explicitly tackling gender to differentiate these categories, something which CIDER could enable, for example, by allowing multiple "gender axes" to be defined.

**3.3.1 Seed words.** The seed words for gender were selected manually. However, the seed word generation methodology implemented in Section 2.3 is not specific to sentiment and can be applied to alternative scales. The seed words used are as follows:

```
male_seeds = ["he", "him", "hes", "his", "himself", "boy", "boys", "dad",
              "dads", "father", "fathers", "brother", "brothers", "gentleman",
              "gentlemen", "male", "males", "man", "men", "masculine", "mr",
              "son", "sons"]

female_seeds = ["girl", "girls", "sister", "sisters", "mom", "moms", "mum",
                "mums", "mother", "mothers", "lady", "ladies", "female",
                "females", "woman", "women", "feminine", "ms", "missus",
                "mrs", "daughter", "daughters", "she", "her", "shes",
                "hers", "herself"]
```

It is worth noting that "miss" and "misses" were intentionally excluded due to their use in other contexts (such as football).

**3.3.2 Generating and filtering polarities.** The method used to apply CIDER for a gender scale is the same as that presented in Section 3.2.2, with two exceptions:

1. When the sentiment polarities were filtered some of the default VADER lexicon remained in the CIDER lexicon. As this is no longer a sentiment classification task we start with an empty lexicon.

2. P5) (the parameter that dictates the percentage of polarised words to return) is set to 0.30 rather than the previous 0.13. This is to counteract the decrease in the base lexicon size. This threshold was manually selected, however, from observation of the returned polarities at different parameter values, 0.30 returned a sufficient lexicon volume whilst maintaining high-quality words.

## 4 Results

### 4.1 Sentiment

A sample of the resulting polarities for the three lexicons is shown in Fig 1. For this figure, the polarities have been scaled between [0, 1], where 0 is negative, and 1 is positive. The size represents the frequency in the dataset, and the colour and location on the x-axis represent the positivity (closer to the right and greener implies more positive). To minimize word overlap in this figure, a force-based repelling algorithm was used to position the words. Because of this, the colour is a slightly more accurate representation of sentiment. However, the impact of this adjustment is minimal, as evidenced by high Pearson's R values between the original and adjusted locations of 0.90 (P<0.001), 0.88 (P<0.001), and 0.96 (P<0.001) for the "hot", "wind", and "cold" datasets respectively. The datasets show a left-skewed bias in polarities, with both a greater number of unique positive words and a greater volume of positive words (similar to the result of Dodds et al. [58]).

Upon filtering the polarities as mentioned in Section 3.2.2, the wind, cold, and hot subsets are classified using their respective custom CIDER classifiers to evaluate the extent to which they agree with the human-annotated labels provided by Asiaee T. et al. [40]. These results are then compared to VADER and the seven additional, unsupervised, general sentiment analysis techniques outlined in Section 3.2.3. The sentiment classification accuracy and weighted F1 score of each approach compared to the human-annotated labels are shown in Table 2.

For accuracy, CIDER performed the highest in two out of the three datasets, and second highest in the third. The average accuracy of CIDER is higher than all other approaches. In particular, the accuracy was significantly better than the default VADER classifier. The same

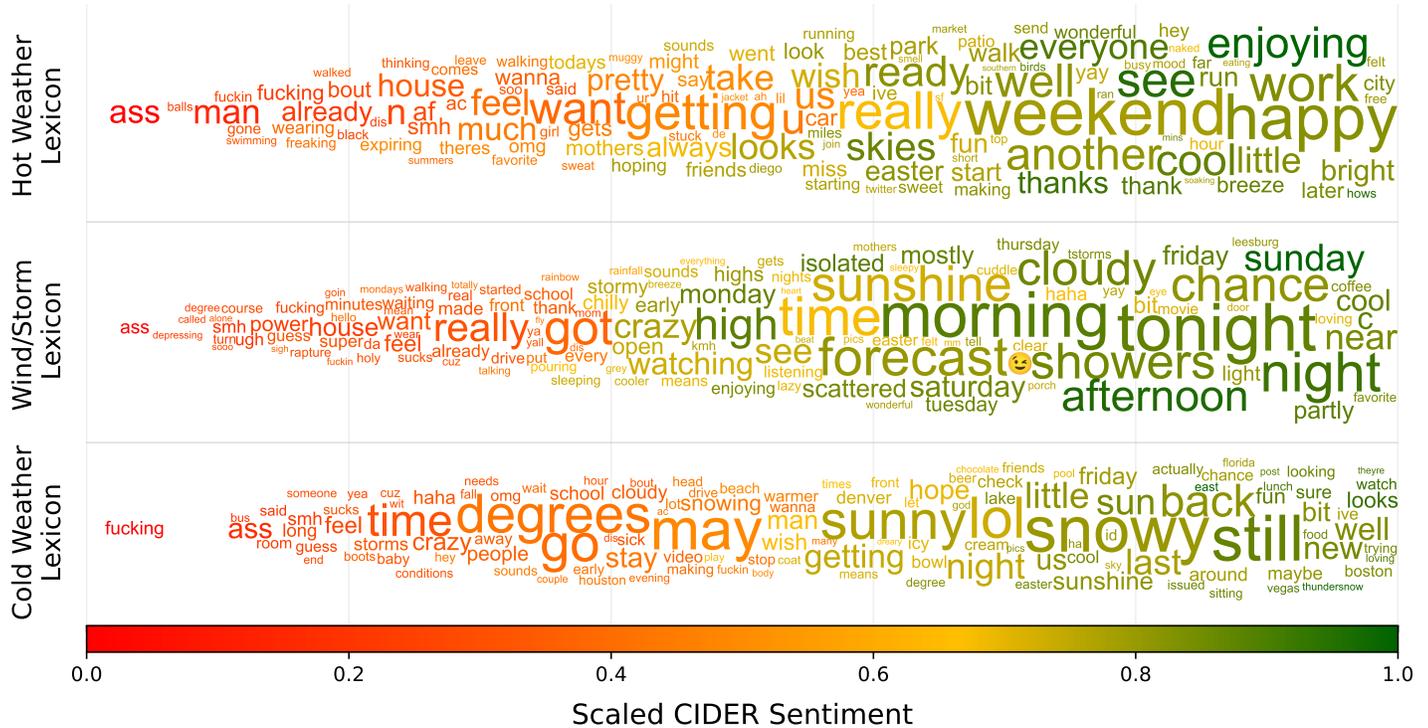

**Fig 1. Sample of CIDER derived sentiment lexicons for the hot weather, wind/storm, and cold weather Twitter datasets.** Both colour and position represent sentiment. Token size represents frequency.

https://doi.org/10.1371/journal.pone.0299490.g001

Table 2. Sentiment classification accuracy and F1 score of individual weather datasets. The highest score in each column is highlighted in green.

| | Accuracy | | | | Weighted F1 | | | |
|---|---|---|---|---|---|---|---|---|
| | Cold | Heat | Wind | Avg. | Cold | Heat | Wind | Avg. |
| **Method** | | | | | | | | |
| **CIDER** | 0.783 | 0.790 | 0.769 | 0.781 | 0.782 | 0.794 | 0.770 | 0.782 |
| **Umigon** | 0.656 | 0.711 | 0.785 | 0.717 | 0.668 | 0.711 | 0.781 | 0.720 |
| **Sentiment140** | 0.671 | 0.694 | 0.720 | 0.695 | 0.690 | 0.730 | 0.713 | 0.711 |
| **SentiStrength** | 0.638 | 0.723 | 0.569 | 0.643 | 0.652 | 0.713 | 0.568 | 0.644 |
| **AFINN** | 0.615 | 0.728 | 0.510 | 0.618 | 0.631 | 0.726 | 0.493 | 0.617 |
| **VADER** | 0.643 | 0.722 | 0.491 | 0.618 | 0.660 | 0.692 | 0.464 | 0.605 |
| **TextBlob** | 0.655 | 0.617 | 0.548 | 0.607 | 0.656 | 0.590 | 0.558 | 0.601 |
| **LIWC15** | 0.564 | 0.710 | 0.520 | 0.598 | 0.572 | 0.697 | 0.515 | 0.594 |
| **LIWC22** | 0.507 | 0.646 | 0.535 | 0.563 | 0.507 | 0.656 | 0.532 | 0.565 |

https://doi.org/10.1371/journal.pone.0299490.t002

applies to the weighted F1 score. A sample of results with language representative of the tweet sample is shown in Fig 2, alongside their scores from CIDER and VADER.

To better understand the disagreement between CIDER and VADER, the individual word polarities can be compared. By applying CIDER to a more expansive corpus than those previously analysed (as detailed in Appendix A in S1 Appendix), we produce a lexicon with greater overlap with the default VADER dictionary, allowing us to assess the agreement between the lexicons, highlighting words which VADER potentially misclassifies. Fig 3 presents the percentile-ranked polarities for both the CIDER and VADER lexicons after CIDER has been trained on the GeoUK dataset. Although a strong positive correlation exists between the lexicons (Pearson's R: 0.743, P < 0.001), notable discrepancies are clear. Specifically, highlighted are the 10 words with the greatest positive difference in rank (VADER classifies as positive, CIDER classifies as negative), and the 10 words with the greatest negative difference in rank (VADER classifies as negative, CIDER classifies as positive). Accompanying this, the adjacent table presents the four words most strongly associated with these 20 highlighted words, as measured by PPMI scores. This provides insights into the potential reasons for VADER's default misclassifications. The majority of these discrepancies are clear.

### 4.2 Alternative scales

In this section, we focused solely on the analysis of the GeoUK dataset. The CIDER model was separately trained twice on this dataset: first with sentiment seed words and then with gender seed words. Whilst sentiment is an intuitive linguistic scale to conceptualise, gender is potentially more abstract. Fig 4, shows example sentences, alongside their gender classification.

| Text | CIDER Score | VADER Score |
|---|---|---|
| "Need AC - way too hot. Take care out there!!" | CIDER_heat: -0.652 (NEG) | VADER: 0.583 (POS) |
| "This storm is super scary. Please pray for us 🙏" | CIDER_wind: -0.633 (NEG) | VADER: 0.649 (POS) |
| "Drinking coffee watching the snow - It can't get better than this!" | CIDER_cold: 0.438 (POS) | VADER: -0.402 (NEG) |

**Fig 2. Text with representative language to the Twitter dataset with their VADER and CIDER classifications.**

https://doi.org/10.1371/journal.pone.0299490.g002

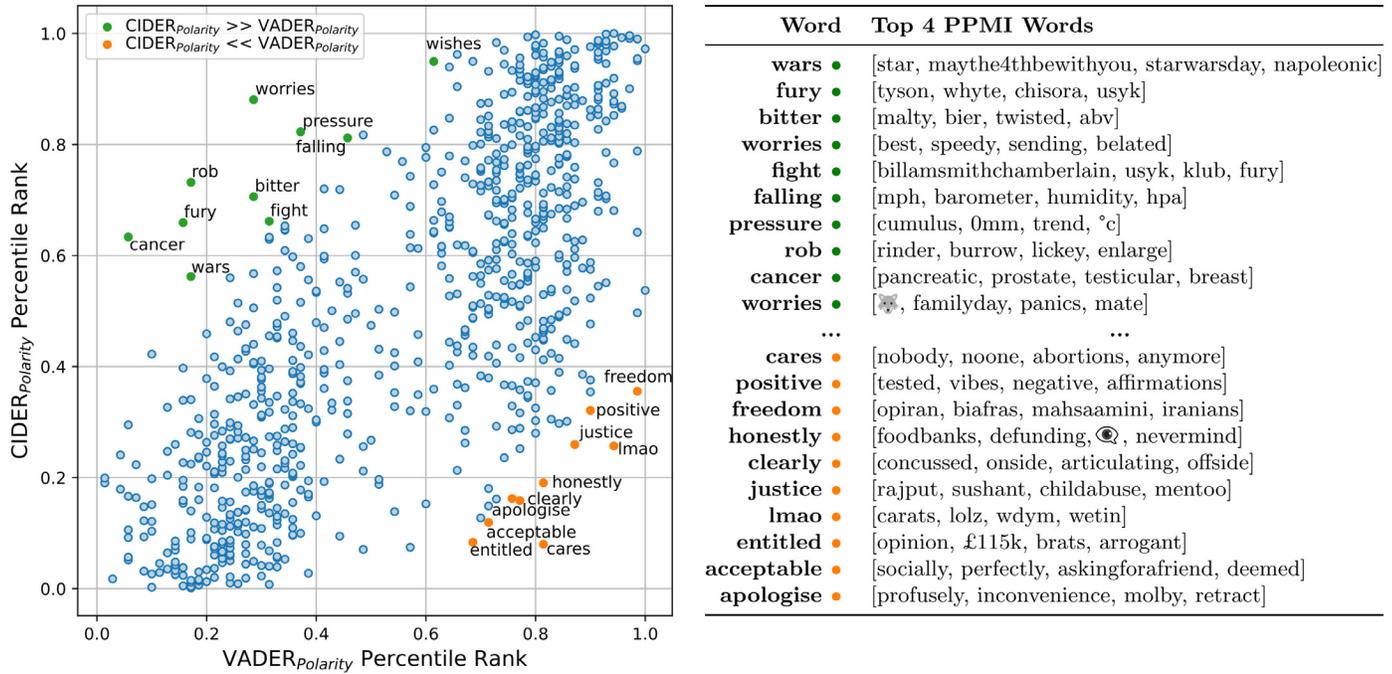

**Fig 3. Sentiment comparison between CIDER lexicon (trained on GeoUK 2022 Tweets) and default VADER lexicon.** Axes are percentile rank polarities, i.e. the lower left quadrant contains words VADER and CIDER have assigned negative polarities to, and the upper right quadrant contains words that VADER and CIDER have assigned positive polarities to.

https://doi.org/10.1371/journal.pone.0299490.g003

To explore the linguistic differences between the generated sentiment and gender lexicons, we investigated the polarities of emojis in the dataset. Emojis provide a rich insight into emotion on social media [59], as well as uncovering regional variations in communication styles [60]. Fig 5 presents the 200 most common emojis in the dataset, with position dictated by their corresponding sentiment and gender lexicon score. Similar to Fig 1, the spatial position of each emoji is not exact. Despite this, both the adjusted gender axis, and adjusted sentiment axis show a strong correlation with CIDER's gender and sentiment results (Pearson's R: 0.81, P<0.0001, and Pearson's R: 0.97, P<0.0001, respectively). Fig 5 shows that the GeoUK 2022 dataset has a clear positive bias, with more positive-female emojis than positive-male emojis,

| Text | Compound | Intensity |
|---|---|---|
| Ronaldo is playing at Old Trafford today | -0.4251 | 0.5161 |
| Out walking my 🐶 and a 🐱 ran towards us | 0.2438 | 0.5168 |
| Enjoyed watching the euros - was a great match 🦁 | 0.8139 | 0.8139 |
| Wish this clown would accept retirement | -0.7804 | 0.7804 |
| Support our school teachers and carers! | 0.4217 | 0.4217 |

**Fig 4. Examples of CIDER classified gendered sentences.** Words classified as male are highlighted in blue, and words classified as female are highlighted in green. Compound scores range from -1 (male) to +1 (female).

https://doi.org/10.1371/journal.pone.0299490.g004

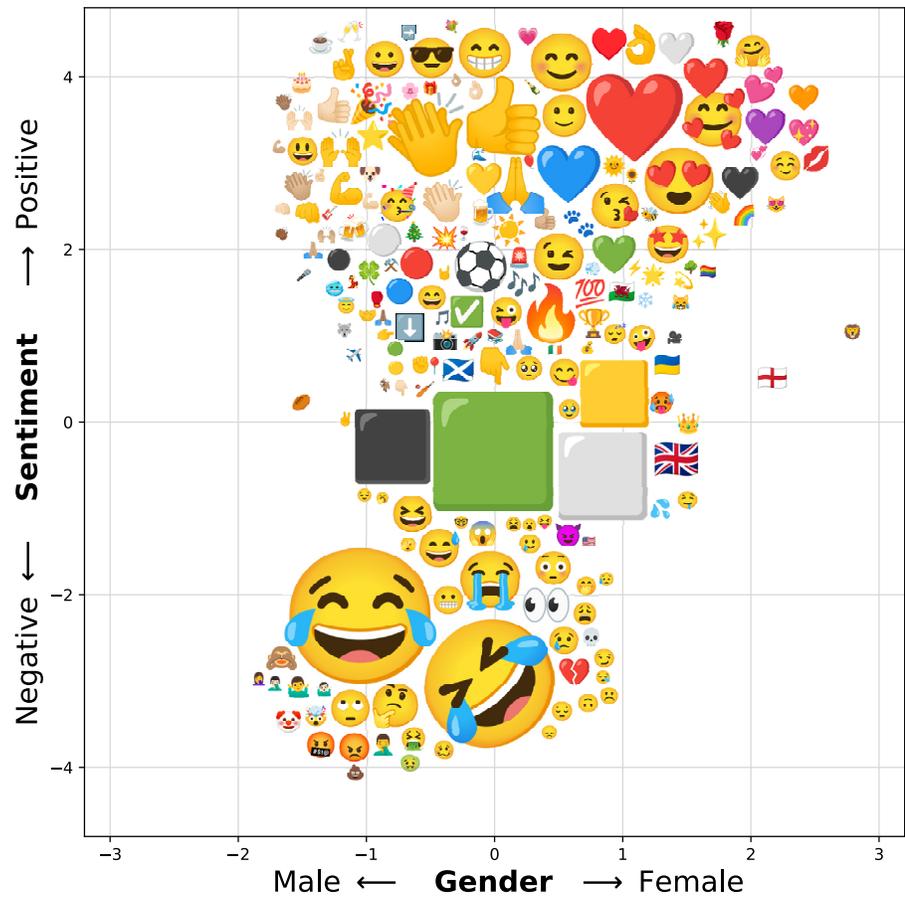

**Fig 5. CIDER gender lexicon plotted against CIDER sentiment lexicon.** Trained on GeoUK 2022 Tweets.

https://doi.org/10.1371/journal.pone.0299490.g005

and more negative-male emojis than negative-female emojis (similar to the findings of Park et al. [61]).

By independently training the CIDER model for gender and sentiment analysis on the GeoUK dataset, we generated distinct classifiers for each dimension. Every tweet was then classified using the models, and their respective intensity metrics were summed to produce an overall intensity score across both two axes. Fig 6 shows the daily mean tweet intensity, after

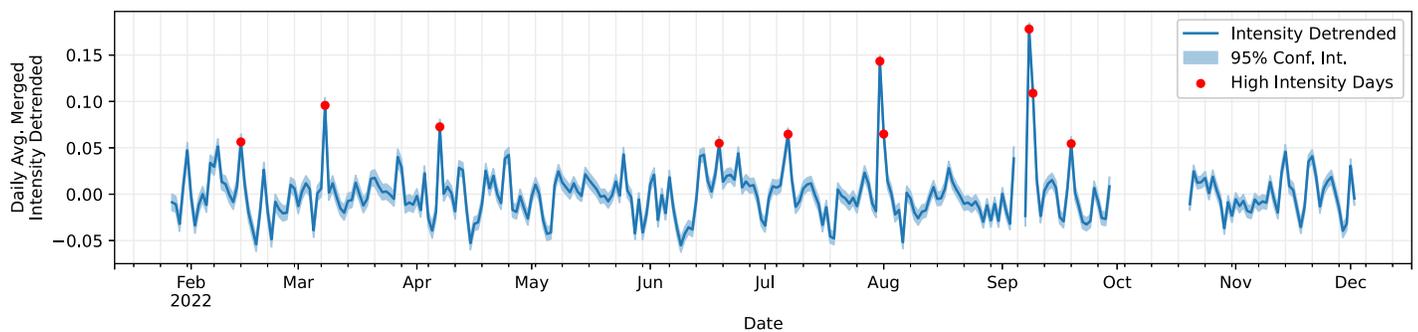

**Fig 6. GeoUK 2022 daily mean tweet intensity (detrended).** Gaps in the timeseries are due to data collection outages. The ten days with the highest intensity have been marked.

https://doi.org/10.1371/journal.pone.0299490.g006

Table 3. Topic modelling results from 10 days with the highest average intensity.

| | Representations | Sentiment | Gender |
|---|---|---|---|
| 14/02/2022 | [valentinesday, happyvalentinesday2022, celebrate, romantic, lovers] | 0.502 | 0.323 |
| 08/03/2022 | [internationalwomensday2022, happyinternationalwomensday, happywomensday, women, womenempowerment] | 0.169 | 0.334 |
| 07/04/2022 | [united, city, liverpool, lads, chelsea] | -0.060 | -0.036 |
| 19/06/2022 | [fathersday, fathersday2022, happyfathersday, dad, happyfathersday2022] | 0.484 | -0.072 |
| 07/07/2022 | [borisjohnson, borisout, borisjohnsonresign, resigned] | -0.268 | -0.183 |
| 31/07/2022 | [lionesseslive, englandwomen, footballs, team, women] | 0.239 | 0.698 |
| 01/08/2022 | [football, womens, englands, mens, sports] | 0.179 | 0.337 |
| 08/09/2022 | [royalfamily, condolences, monarch, queenelizabeth, saddened] | -0.220 | 0.607 |
| 09/09/2022 | [queen, royal, majesty, queens, monarchy] | -0.107 | 0.376 |
| 19/09/2022 | [funeral, mourning, coffin, death, queen] | -0.240 | 0.458 |

The five most representative tokens for each day's dominant topic (as determined by Bertopic) alongside gender scores ranging from -1 (male) to +1 (female), and sentiment scores ranging from -1 (negative) to +1 (positive).

https://doi.org/10.1371/journal.pone.0299490.t003

detrending using SciPy's "seasonal detrend" function [62]. The ten days with the highest average intensity have been highlighted.

To further investigate the content posted on these high-intensity days, Bertopic, a BERT-based topic modelling package, was used [63]. This tool clusters the tweets into discrete thematic categories. Table 3 shows the five most representative words for each highlighted day's dominant topic. The mean gender and sentiment scores for the tweets assigned to the daily dominant category have been calculated. These scores have been normalised by the average sentiment and the average gender tweet polarity respectively. The results are largely intuitive, with each day's most common topic being a clearly identifiable event of 2022.

## 5 Discussion

This study introduces a new approach to sentiment analysis that takes context into account. By combining the dictionaries generated by SocialSent with the scoring rules and base sentiment dictionary of VADER, CIDER outperforms state-of-the-art unsupervised sentiment analysis methods. Our implementation is fast; it takes about 30 minutes to train on 30 million tweets and requires minimal supervision. The model only needs a small set of around 20 seed words, which can be specified manually or through our implemented seed word selection algorithm. While more advanced large language models (LLMs) like BERT and GPT can be used for sentiment analysis and may yield better accuracy, CIDER has several clear advantages:

- Transparency. This is crucial for interpretability, especially in fields where understanding the reasoning behind a model's decision is important. In contrast, LLMs and most supervised classification algorithms are not easily explained and often operate as "black boxes", making it challenging to understand the nuances of the sentiment analysis i.e. the reason why a particular sentence gets the score it does. This transparency also facilitates easier compliance with data privacy and ethical guidelines, a growing concern in NLP applications. In contrast, the opaque nature of LLMs can make it challenging to ensure such compliance.

- Classification speed improvements. CIDER's sentiment analysis employs a computationally efficient dictionary-based indexing approach. This allows for rapid classification and requires very little memory.

- Training advantages. LLMs like BERT require extensive training time and computational resources, often needing specialised hardware like high-performance GPUs. Fine-tuning BERT for cross-domain sentiment analysis would worsen this issue, making the process both time-consuming and resource-intensive. In contrast, CIDER offers a streamlined training process that is both time-efficient and memory-efficient. This makes CIDER an accessible option for a broad range of domains and disciplines, especially those with limited high-performance computing resources. Moreover, CIDER's efficiency in training is particularly advantageous in scenarios where rapid deployment or frequent model updates are required, a context where LLMs, due to their size and complexity, might be less feasible.

- Data-availability. As CIDER is unsupervised, it eliminates the need for costly and time-consuming labelling of text for each new context. Moreover, for fine-tuning an LLM to scales other than sentiment it can be challenging to manually categorise text to create training data without full domain context.

CIDER's applicability extends beyond sentiment analysis, as demonstrated in our gender case study. Its flexible framework allows for the exploration and classification of a variety of scales. This adaptability makes CIDER a valuable tool for interdisciplinary research, including gender studies, by providing a more accessible method for linguistic analysis. This democratises NLP and linguistic research, enabling researchers from academic disciplines other than computer science to make use of NLP tools. By making a Python package which is easy to install and run we hope that CIDER can find widespread use. Furthermore, CIDER enables multi-dimensional analysis, allowing for studies across multiple scales. For example, it could be applied to a subjective/objective scale using seed words such as ["feel", "believe", "think"] and ["fact", "know", and "prove"], to differentiate how subjective-negative language varies from subjective-positive language. Other possible applications include political left-right axes, or geographic axes (east-west, north-south), which could be useful for computational linguistic applications [64].

Moving forward, further research into how CIDER can be optimised for small datasets should be carried out. One potential approach is to improve seed word selection. For smaller datasets, variation in the initial seed word choices can have a substantial impact on the final produced lexicon. To improve CIDER's effectiveness on these datasets, extracting the most robust initial set of seed words is important. An additional area of investigation for small datasets is the use of pre-trained embeddings which can be optimised for the smaller dataset using the generated PPMI matrix. Another area of development we are investigating is incorporating named entities into CIDER. Currently, words are tokenised into unigrams, however, by applying named entity recognition (for instance spaCy [65]), important entities can be preserved in the text, providing more context to the final lexicon. Finally, research into alternative scales is difficult to validate, for instance, the true "gender" of a sentence. By employing a diverse selection of participants from the UK to manually label a sample of tweets as masculine and feminine, CIDER could be validated for gender as well as sentiment.

In the future, this method can allow for temporal tracking of public sentiment, helping monitor fluctuations within specific domains such as discussions on climate change or national elections. CIDER could also be extended to multiple languages: the training step only requires a small number of seed words which could easily be gathered from fluent speakers. The mutation rules and base sentiment dictionary of VADER are specific to English but could in principle be modified to the target language. Compared to manually tagging training data this approach may require a less nuanced understanding of the language, e.g. sarcastic tweets are detected by CIDER without any specific training examples of sarcasm. It also only needs to be done once for each language. This would facilitate the study of emotional responses and sentiment variations across diverse linguistic and cultural contexts.

This research demonstrates that CIDER can overcome the limitations of existing sentiment analysis approaches for large collections of short texts about a particular theme or topic. Moreover, it provides an easy-to-use tool to investigate linguistic scales beyond sentiment. We hope that the implementation provided is a useful package for researchers interested in linguistic analysis.

## Supporting information

**S1 Appendix.**
(PDF)

**S1 File.**
(ZIP)

**S2 File.**
(7Z)

## Acknowledgments

All authors have read and agreed to the published version of the manuscript.

## Author Contributions

**Conceptualization:** James C. Young, Rudy Arthur.

**Data curation:** James C. Young.

**Formal analysis:** James C. Young.

**Investigation:** James C. Young.

**Methodology:** James C. Young.

**Software:** James C. Young.

**Supervision:** Rudy Arthur, Hywel T. P. Williams.

**Validation:** James C. Young.

**Visualization:** James C. Young.

**Writing – original draft:** James C. Young.

**Writing – review & editing:** Rudy Arthur, Hywel T. P. Williams.